\documentclass[10pt,twocolumn,letterpaper]{article}

\usepackage{iccv}
\usepackage{times}
\usepackage{epsfig}
\usepackage{graphicx}
\usepackage{amsmath}
\usepackage{amssymb}

\usepackage{amsthm}
\newtheorem{mythm}{Theorem}
\usepackage{multirow}
\usepackage{caption}
\usepackage{subcaption}
\usepackage{soul}
 \usepackage{color} 

\usepackage[pagebackref=true,breaklinks=true,letterpaper=true,colorlinks,bookmarks=false]{hyperref}

\renewcommand{\vec}[1]{\mathbf{#1}}
\newcommand{\argmin}{\mathop{ \arg\!\min}}

\def\lt{\left}
\def\rt{\right}
\def\x{\vec{x}}
\def\y{\vec{y}}
\def\z{\vec{z}}
\def\k{\vec{k}}

\def\w{\vec{w}}
\def\t{\vec{t}}
\def\c{\vec{c}}

\iccvfinalcopy 

\def\iccvPaperID{1106} 

\ificcvfinal\fi
\begin{document}

\title{Improving Image Restoration with Soft-Rounding}

\author{Xing Mei$^{1,3}$ \qquad Honggang Qi$^{2}$ \qquad Bao-Gang Hu$^{3}$ \qquad Siwei Lyu$^{1}$\\
$^1$Computer Science Department, University at Albany, SUNY\\
$^2$Computer Science Department, University of Chinese Academy of Sciences\\
$^3$NLPR, Institute of Automation, Chinese Academy of Sciences, Beijing, China\\
{\tt\small~xmei@albany.edu,~hgqi@jdl.ac.cn,~hubg@nlpr.ia.ac.cn,~slyu@albany.edu}
}

\maketitle

\begin{abstract}
Several important classes of images such as text, barcode and pattern images have the property that pixels can only take a distinct subset of values. This knowledge can benefit the restoration of such images, but it has not been widely considered in current restoration methods. In this work, we describe an effective and efficient approach to incorporate the knowledge of distinct pixel values of the pristine images into the general regularized least squares restoration framework. We introduce a new regularizer that attains zero at the designated pixel values and becomes a quadratic penalty function in the intervals between them. When incorporated into the regularized least squares restoration framework, this regularizer leads to a simple and efficient step that resembles and extends the rounding operation, which we term as {\em soft-rounding}. We apply the soft-rounding enhanced solution to the restoration of binary text/barcode images and pattern images with multiple distinct pixel values. Experimental results show that soft-rounding enhanced restoration methods achieve significant improvement in both visual quality and quantitative measures (PSNR and SSIM). Furthermore, we show that this regularizer can also benefit the restoration of general natural images.
\end{abstract}

\section{Introduction}
\label{sec:1}
The task of image restoration is to recover an image from its noisy and/or blurry observation~\cite{Banham:1997:DIR}. Image restoration is an example of ill-posed inverse problems, and unique solutions can only be obtained by introducing proper regularizations and constraints on the clean images. Indeed, it is now generally accepted that the more prior knowledge we have about the properties of the uncorrupted images, the better we can solve the image restoration problem~\cite{Afonso:2010:ALM,Heide:2014:FlexISP}.

To date, the most widely used statistical regularities for images are obtained from their band-pass filter responses (such as gradients or wavelets) or mean-removed pixel patches. Such responses have been observed to have heavy-tailed marginal histograms, on the basis of which regularizers or priors that prefer sparse responses have been used to attain the state-of-the-art restoration performances. Examples include regularizers in the form of the $\ell_0$ norm \cite{Xu:2011:ISV,Xu:2013:UL0} and total variation~\cite{Rudin:1992:TV}, or prior models based on generalized Laplacian \cite{Krishnan:2009:FID,Levin:2007:IDC} and Gaussian mixtures \cite{Wainwright:2000:GSM,Zoran:2011:EPLL}. In these image restoration methods, pixel values are usually treated as continuous variables.

\begin{figure*}[t]
\begin{center}
\begin{subfigure}[t]{.19\textwidth}
\centering
\includegraphics[width=\textwidth]{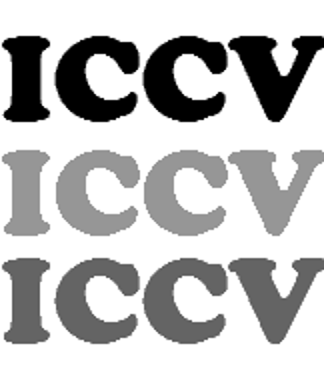}
\caption{\small original image}
\end{subfigure}
\begin{subfigure}[t]{.19\textwidth}
\centering
\includegraphics[width=\textwidth]{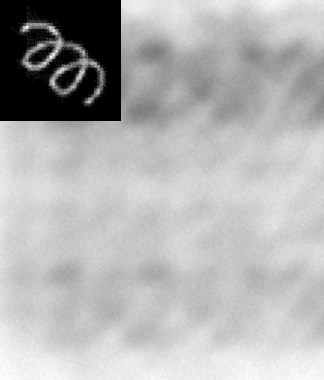}
\caption{\small corrupted image}
\end{subfigure}
\begin{subfigure}[t]{.19\textwidth}
\centering
\includegraphics[width=\textwidth]{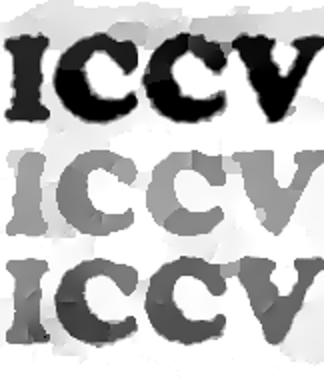}
\caption{\small \centering $\ell_0$ restoration~\cite{Pan:2014:DTI} \newline PSNR = 19.45dB \newline  SSIM = 0.79}
\end{subfigure}
\begin{subfigure}[t]{.19\textwidth}
\centering
\includegraphics[width=\textwidth]{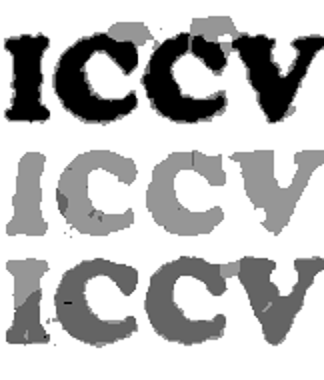}
\caption{\small \centering $\ell_0$ + post rounding \newline PSNR = 19.28dB \newline SSIM = 0.82}
\end{subfigure}
\begin{subfigure}[t]{.19\textwidth}
\centering
\includegraphics[width=\textwidth]{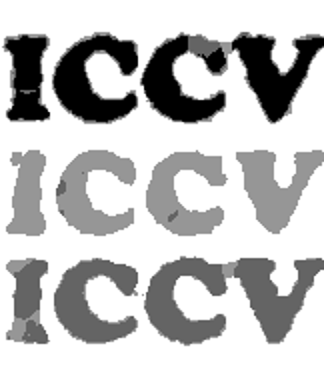}
\caption{\small \centering $\ell_0$ + soft-rounding \textbf{\newline PSNR = 19.84dB \newline SSIM = 0.85}}
\end{subfigure}
\end{center}
~\vspace{-2em}
\caption{\em \small An example demonstrating the effect of our method. {\bf (a)} An original image with four distinct pixel values $0, 100, 150, 255$. {\bf (b)} Its corruption with a $101 \times 101$ blur kernel and $1\%$ white Gaussian noise. {\bf (c)} Restoration results using the state-of-the-art method based on $\ell_0$ regularization of image gradients~\cite{Pan:2014:DTI}. The artifacts in the restored image are due to pixel values that are not consistent with the four distinct pixel values. {\bf (d)} Rounding (nearest neighbor) of the restoration results in (c) to the four distinct pixel values. Though some artifacts in (c) are removed, some structures restored in (c) are also destroyed. {\bf (e)} Restoration results using our method that combines gradient domain $\ell_0$ regularization based restoration with soft-rounding operation. Note that our method achieves significant improvement in comparison with (c) and (d) both visually and quantitatively (PSNR and SSIM).}
\label{fig:teaser}
~\vspace{-2.5em}
\end{figure*}

However, images are captured and stored in a digital format, which poses a general constraint on pixel values. The pixels of an image in a $b$-bit format can only take integer values from the set $\{ 0, \cdots, 2^b-1\}$. For some important subclasses of images such as text or pattern images, their pixels can take only a distinct set of the full range integer values (see Fig.\ref{fig:teaser}(a) for an example).  Furthermore, it is usually possible to obtain these values before we restore such an image. For text and barcode images, the pixel values are usually separated into two distinct classes of foreground pixels and background pixels. For pattern images, image pixels usually take multiple distinct pixel values, which can be estimated from other images of the same type.

Knowing that the pristine image can only take some distinct pixel values provides a useful constraint for image restoration, which can help to suppress visual artifacts and improve the restoration performance. A straightforward approach would be to round the restored pixel values to the nearest distinct pixel values as a post-processing step. However, such a simple solution often does more harm than good, as the rounding step can undo the structures recovered from the restoration step (see Fig.\ref{fig:teaser}(c)). On the other hand, except in a few cases~\cite{Joshi:2009:IDD,Mei:2015:UniHIST,Pan:2014:DTI,Zhang:2012:AAM}, such prior information about pixel values seems trivial and is largely overlooked by most existing image restoration algorithms.

In this work, we describe an effective and efficient approach to incorporate the knowledge of distinct pixel values of the pristine images into the popular regularized least squares restoration framework. We introduce a new regularizer of the uncorrupted images in the pixel domain. This regularizer attains zero at the designated pixel values and is a quadratic function in the intervals between them. As such, it reflects the preference of the pixel values of the restored image to take the designated distinct pixel values.  When incorporated into the regularized least squares restoration framework, this regularizer leads to a simple and efficient solution that resembles the rounding operation, which we term as {\em soft-rounding}. We apply the soft-rounding enhanced solution to various state-of-the-art image restoration methods, and show that it achieves significant improvement in both visual quality and quantitative measures (PSNR and SSIM) for the restoration of binary images and multi-value pattern images. We also demonstrate its effectiveness for general natural image restoration.

To sum up, the main contributions of this work are:
~\vspace{-.5em}
\begin{itemize} \itemsep -0.2em
\item we introduce a new regularizer encouraging the output image to take designated distinct pixel values;
\item unlike previous binary image restoration methods~\cite{Pan:2014:DTI,Zhang:2012:AAM}, our regularizer can intrinsically handle more than two distinct pixel values;
\item optimization with this regularizer leads to a simple and efficient soft-rounding operation that allows easy integration with existing restoration algorithms;
\item when applied to the restoration of text and pattern images, our method achieves significant improvement in both quantitative and visual quality.
\end{itemize} 

\section{Background and Related Work}
\label{sec:2}
We formulate the image restoration problem as recovering a pristine image of $m$ pixels, which is vectorized and denoted with an $m$-dimensional vector $\x$, from its noisy and/or blurry observation $\y$. We assume that $\y$ is generated from the convolution of $\x$ with a spatially invariant blurring kernel $\k$ and further contaminated with additive Gaussian noise $\mathbf{n}$, as:
~\vspace{-.245em}
\begin{equation} \textstyle
\y = \k \otimes \x + \vec{n},
\label{eq:1}
~\vspace{-.245em}
\end{equation}
where $\otimes$ denotes the convolution operation~\cite{Banham:1997:DIR}\footnote{In the current work, we consider the setting that the blurring kernel and noise variance are known, known as {\em non-blind} restoration. Our method can also be extended to blind restoration~\cite{Levin:2009:BDA}.}. The restoration problem can be solved within a {\em regularized least squares} (RLS) estimation framework, as:
\begin{equation}
\min\limits_{\x} \frac{1}{2}\| \y-K\x \|^2_2+\lambda_{N}\Gamma_N(D\x)
\label{eq:2}
\end{equation}
where $K$ corresponds to the block Toeplitz matrix representation of the convolution kernel $\k$, $D$ represents a linear transform through which image properties can be better modeled, $\Gamma_N$ is a regularizer in the transformed domain, and $\lambda_{N}$ is a adjustable parameter balancing the contribution of the data fidelity term and the regularization term in the objective function. For image denoising, $K$ reduces to an identity matrix.

The choice of the linear transform $D$ and the regularizer $\Gamma_N$ is essential for effective restoration. A common approach is to first transform image to band-pass domains (\eg, gradients or wavelets) or use mean-removed patches, then use regularizers in the form of sparsity-encouraging norms, \eg, $\ell_0$~\cite{Xu:2011:ISV,Xu:2013:UL0} and total variation~\cite{Rudin:1992:TV}, or models capturing statistical properties of images in such domains, \eg, generalized Laplacian~\cite{Krishnan:2009:FID,Levin:2007:IDC}, Gaussian scale mixtures \cite{Wainwright:2000:GSM}, or general Gaussian mixtures~\cite{Zoran:2011:EPLL}. 


To date, properties of the pristine images in the pixel domain have been largely overlooked in image restoration but in a few cases~\cite{Chen:2011:EDI,Joshi:2009:IDD,Mei:2015:UniHIST,Pan:2014:DTI,Zhang:2012:AAM}. The constraint on pixels to take distinct values has been considered in a few recent works in binary image restoration, where the pixels can take only two values. Zhang introduced a regularization term to penalize pixel values that drift away from the two values~\cite{Zhang:2012:AAM}. The specific regularizer used there is a fourth-order polynomial and requires iterative numerical approximations on each pixel. Pan \etal used a $\ell_0$ regularization term if only the zero peak in pixel values is considered~\cite{Pan:2014:DTI}. They showed that this $\ell_0$ intensity prior helps to identify salient edges in text images and therefore improves blur kernel estimation, but its non-blind restoration performance was not reported in the paper. A potential limitation of the $\ell_0$ intensity prior is that it might not be suitable for binary images whose pixel values center around two peaks far away from zero. Another common limitation of these binary image restoration methods is that they can not be easily extended to handle more than two distinct pixel values.

Some recent methods further incorporate histograms of the pixel values in image restoration. Chen \etal studied the intensity histograms of clear document images and used them as reference information for blind deconvolution, leading to improved restored results with reduced visual artifacts~\cite{Chen:2011:EDI}. Mei \etal recently proposed a framework to enforce marginal histogram constraints in image restoration. They showed that for pattern images, marginal intensity histograms can significantly improve the denoising performance under high noisy levels~\cite{Mei:2015:UniHIST}. Although a pixel histogram provides more information than specifying distinct pixel values in an image, estimating such histograms from degraded images is a non-trivial task~\cite{Cho:2010:CAP,Mei:2015:UniHIST,Wangmeng:2014:GHP}, which poses a limitation for these methods in practice. In contrast, as we pointed out in Section \ref{sec:1}, obtaining distinct pixel values before restoring an image is usually possible for the type of images we focus on in this work.


\section{Method}
\label{sec:3}
As discussed previously, some subclasses of images have distinct pixel values. When these values are known {\em a priori}, they provide valuable information to improve the restoration performance. In this section, we describe in detail an approach to incorporate such information into the general RLS restoration framework, Eq.\eqref{eq:2}. It turns out that the specific solution we obtained has a particularly intuitive interpretation as a ``soft-rounding'' operation, which adaptively adjust the restored pixel values to the target values.

\subsection{Distinct Pixel Value Regularizer}
\label{sec:31}

We first introduce a regularizer to encourage pixel values of the restored image to take the designated distinct pixel values. Formally, assuming we require the restored image should have $n$ distinct pixel values $t_1 < t_2 \cdots < t_n$, represented in a vector $\t = (t_1, t_2, \cdots, t_n)$, we define the regularizer as
\begin{equation}
\textstyle \Gamma_I(\x)=\sum_{i=1}^m \gamma_\t(x_i)
\label{eq:3}
\end{equation}
where $m$ is the number of the image pixels, and $\gamma_\t$ measures the intensity deviation on each pixel:
\begin{equation}
\gamma_\t(x) =
\begin{cases}
\frac{1}{2}(t_1-x) & \mbox{if $x<t_1$}\\
\frac{1}{2}(x-t_j)(t_{j+1}-x) & \mbox{if $x\in[t_j,t_{j+1}]$}\\
\frac{1}{2}(x-t_n) & \mbox{if $x>t_{n}$}
\end{cases}.
\label{eq:4}
\end{equation}


\begin{figure}[t]
\begin{center}
\begin{tabular}{c}
\includegraphics[width=0.25\textwidth]{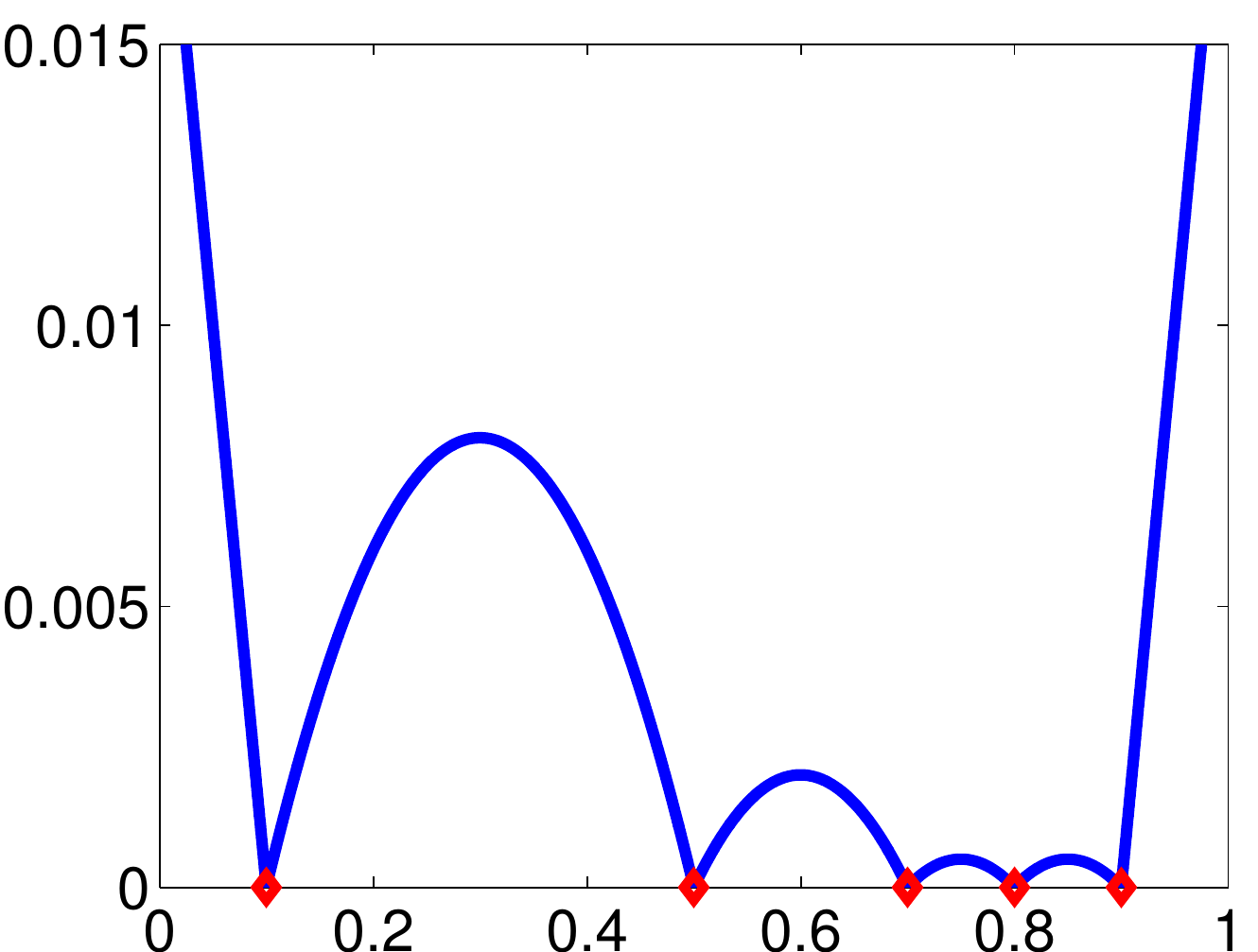}
\end{tabular}
\end{center}
~\vspace{-2em}
\caption{\em \small An example of $\gamma_\t(x)$ in the range of $[0,1]$, with five distinct values (\textcolor{red}{$\diamond$}) corresponding to $[0.1, 0.5, 0.7, 0.8, 0.9]$.}
\label{fig:2}
~\vspace{-2.5em}
\end{figure}

Fig.\ref{fig:2} shows an example of $\gamma_\t(x)$ in the range of $[0,1]$.  This function has the following distinct characteristics:
~\vspace{-.5em}
\begin{enumerate} \itemsep -.2em
\item $\gamma_\t(x) \geq 0$ for $x \in {\mathcal R}$;
\item $\gamma_\t(x)$ attains zero at $t_j~(j=1,\cdots,n)$ and reaches local maximum when $x$ locates at the midpoint of each interval;
\item for $x \in [t_j,t_{j+1}]$, $\gamma_\t(x)$ is a concave quadratic function, and $\gamma_\t(x)$ reduces to a linear function outside the range of $[t_1, t_n]$.
~\vspace{-.5em}
\end{enumerate}
Note that $\gamma_\t(x)$ is neither a convex nor a concave function over the whole real line, but at individual intervals it is concave (it becomes either a quadratic or linear function).

\subsection{Using $\Gamma_I$ in Image Restoration}
\label{sec:32}
We incorporate regularizer $\Gamma_I(\x)$ into the RLS restoration framework and solve the following problem:
\begin{equation}
\min\limits_{\x} \frac{1}{2}\| \y-K\x \|^2_2+\lambda_{N}\Gamma_N(D\x)+{\lambda_I\Gamma_I(\x)}
\label{eq:5}
\end{equation}
where we introduce two parameters $\lambda_N,\lambda_I$ to balance the contributions of the three terms in the overall objective function. Note that Eq.\eqref{eq:5} is a non-convex problem, and we optimize it for a local minimum.

Following the general variable splitting scheme~\cite{Afonso:2010:VS}, we introduce an auxiliary variable $\z_I$ and convert Eq.\eqref{eq:5} to the following equivalent problem:
\begin{equation}
\begin{array}{ccl}
& \min\limits_{\x} & \frac{1}{2}\| \y-K\x \|^2_2+\lambda_{N}\Gamma_N(D\x)+{\lambda_I\Gamma_I(\z_I)}\\
&\text{s.t.} & \z_I = \x
\end{array}
\label{eq:6}
\end{equation}
We then solve this equality constrained problem using the {\em augmented Lagrangian method} (ALM)~\cite{Nocedal:2006:NO}. Introducing a Lagrange multiplier $\w_I$ to the equality constraint and a penalty parameter $\mu_I$, we form the augmented Lagrangian function of problem \eqref{eq:6}, as
\begin{equation}
\begin{array}{lll}
& \mathcal{L}(\x,\z_I,\w_I) = \frac{1}{2}\| \y-K\x \|^2_2+\lambda_{N}\Gamma_N(D\x)\\
& +{\lambda_I\Gamma_I(\z_I)} + \frac{\mu_I}{2}\| \x-\z_I \|^2_2 - \w_I^T(\x-\z_I)
\end{array}
\label{eq:7}
\end{equation}
Starting with initial values for $\z_I,\x,\w_I$ and parameter $\mu_I$, we iteratively update $\z_I,\x,\w_I$ with regards to the augmented Lagrangian starting from $k =1$ as follows:
~\vspace{-.5em}
\begin{enumerate} \itemsep -.2em
\item Update $\z_I$: $\z_I^{k+1} = \argmin_{\z_I}\mathcal{L}(\x^{k},\z_I,\w_I^{k})$
\item Update $\x$: $\x^{k+1}  = \argmin_{\x}\mathcal{L}(\x,\z_I^{k+1},\w_I^{k})$
\item Update $\w_I$: $\w_I^{k+1}  = \w_I^{k}-\mu_I(\x^{k+1}-\z_I^{k+1})$.
~\vspace{-.5em}
\end{enumerate}
It is guaranteed in theory that the solution to the original problem \eqref{eq:6} can be obtained by solving \eqref{eq:7} iteratively in the ALM framework when penalty parameter $\mu_I$ increases to a sufficiently large value~\cite{Nocedal:2006:NO}. Compared with the half-quadratic splitting method {\cite{Geman:1995:HQ,Krishnan:2009:FID}}, a significant advantage of ALM is that its convergence can be assured without increasing $\mu_I$ indefinitely, which is especially suitable for solving problems with multiple equality constraints.

The update step for the multiplier (step 3) is straightforward as required by the ALM framework~\cite{Nocedal:2006:NO}. The subproblem updating $\x$ (step 2) can be reformulated as a RLS problem:
\begin{equation}
\min\limits_{\x} \frac{1}{2}\left\| \y-K\x \right\|^2_2 + \frac{\mu_I}{2}\left\|\z_I^{k+1}+\frac{\w_I^{k}}{\mu_I} - \x\right\|_2^2 + \lambda_N\Gamma_N(D\x)
\label{eq:8}
\end{equation}
Compare to Eq.\eqref{eq:2}, Eq.\eqref{eq:8} has an additional quadratic term about $\x$, \ie, the second term. If $\Gamma_N$ is also a quadratic term on $\x$, this subproblem can be solved through conjugate gradient or FFT-based method. For most existing restoration methods with non-smooth $\Gamma_N$ regularizers, this subproblem can be further decomposed into smaller subproblems through half-quadratic splitting or ALM~\cite{Geman:1995:HQ,Krishnan:2009:FID}, for which well-developed proximal operators for existing $\Gamma_N$ regularizers can be reused in the computation process. Note that for practical implementations, the decomposition of the $\x$ subproblem can be done within the same ALM framework as $\z_I$, and its solution requires only a one-iteration approximate updating process. Compared to standard restoration methods with only $\Gamma_N$ terms, our method requires extra slight computation costs on the $\z_I$ and $\w_I$ updating steps. For better understanding of the $\x$ subproblem, please refer to the supplemental material for two concrete examples involving different $\Gamma_N$ terms. The solution of the $\z_I$ (step 1) is detailed in the following section.

\subsection{Soft-Rounding Operator}
\label{sec:33}

After rearranging terms, the subproblem of updating $\z_I$ (step 1) reduces to the following minimization problem:
\begin{equation}
\min\limits_{\z_I} \frac{\mu_I}{2}\lt\|\z_I-\c\rt\|_2^2+\lambda_I\Gamma_I(\z_I),
\label{eq:9}
\end{equation}
where we define $\c = \x^{k}-\frac{1}{\mu_I}\w_I^{k}$. Note that both terms in Eq.\eqref{eq:9} can be split into the the sum on each element of $\z_I$ (see the definition of $\Gamma_I(\cdot)$ in Eq.\eqref{eq:3}). Therefore this problem can be optimized separately on each element as
\begin{equation}
(\z_I^{k+1})_i = \phi_{\t,\frac{\lambda_I}{\mu_I}}(c_i),
\label{eq:10}
\end{equation}
where
\begin{equation}
\phi_{\t,\lambda}(c) = \argmin_x \left(\frac{1}{2\lambda}(x-c)^2 + \gamma_\t(x)\right)
\label{eq:11}
\end{equation}
can be seen as the {\em proximal operator} of function $\gamma_\t(\cdot)$~\cite{Parikh:2013:PA}.

The following result shows that $\phi_{\t,\lambda}(c)$ can be computed with a piece-wise function.
~\vspace{-.5em}
\begin{mythm}
For constant $\lambda > 0$ and $t_1 < t_2 \cdots < t_n$, the optimal solution to Eq.\eqref{eq:11} is given by:
\smallskip

\noindent (i) for $c < t_1$, $\phi_{\t,\lambda}(c) = \min(t_1,c+\frac{\lambda}{2}$);
\smallskip

\noindent (ii)  for $c > t_n$, $\phi_{\t,\lambda}(c) = \max(t_n,c-\frac{\lambda}{2})$;
\smallskip

\noindent (iii) hard-rounding: for $c \in [t_1,t_n]$ and $\lambda \geq 1$,
\begin{equation}
\phi_{\t,\lambda}(c) = %
\begin{cases}
t_j & c \in [t_j, t_j+ d_j ]\\
t_{j+1} & c \in ( t_{j+1} - d_j, t_{j+1} ]
\end{cases}
\label{eq:12}
\end{equation}
where $d_j =  \frac{1}{2} (t_{j+1} - t_j)$ and $j = 1, \cdots, n-1$;

\noindent (iv) soft-rounding: for $c \in [t_1,t_n]$ and $\lambda \in (0,1)$,
\begin{equation}
\phi_{\t,\lambda}(c) = %
\begin{cases}
t_j & c \in [t_j, t_j+ d_j ]\\
 \frac{c}{1 - \lambda} - \frac{\lambda( t_j + t_{j+1})}{2(1 - \lambda)} & c \in [t_j+ d_j ,   t_{j+1} - d_j ]\\
t_{j+1} & c \in [   t_{j+1} - d_j, t_{j+1} ]
\end{cases}
\label{eq:13}
\end{equation}
where $d_j =  \frac{\lambda}{2} ( t_{j+1} - t_j)$ and $j = 1, \cdots, n-1$.
\smallskip
\label{thm}
~\vspace{-.5em}
\end{mythm}


\begin{figure}[t]
\begin{center}
\includegraphics[width=0.23\textwidth]{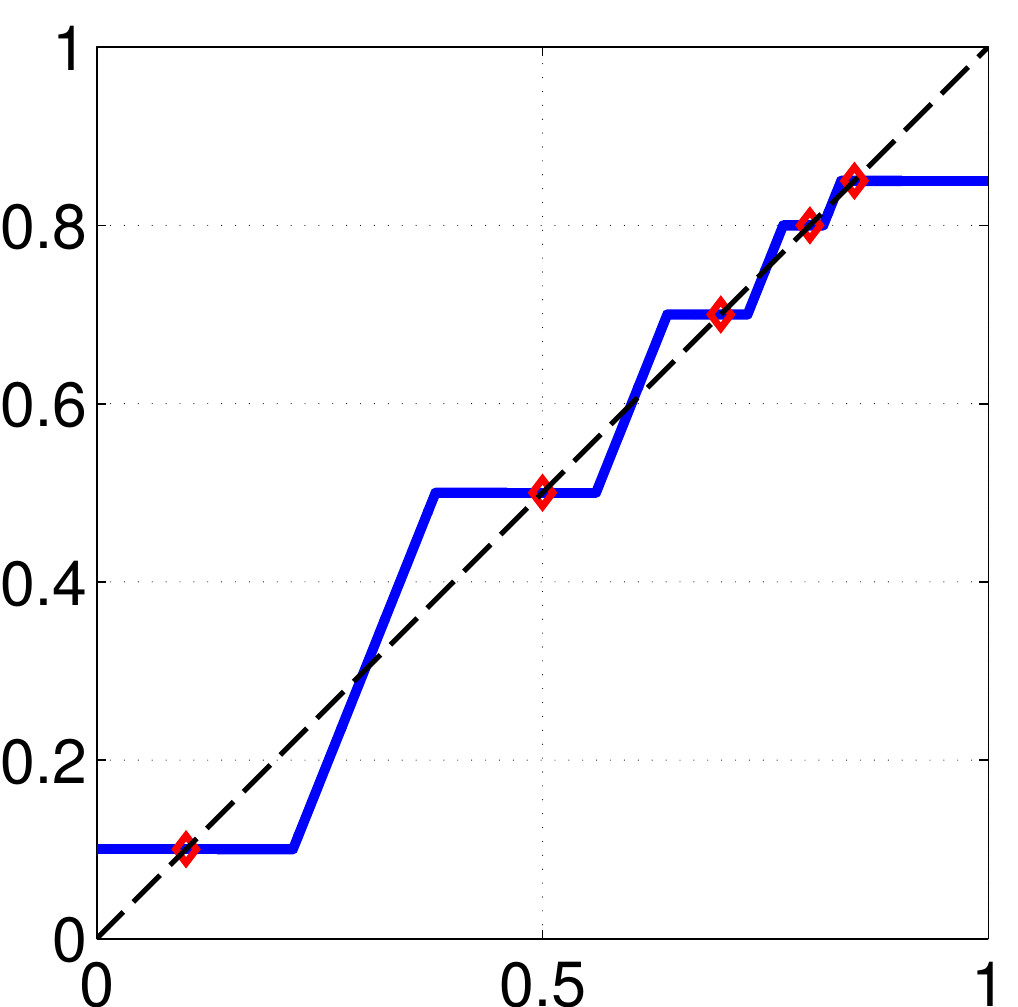}
\includegraphics[width=0.23\textwidth]{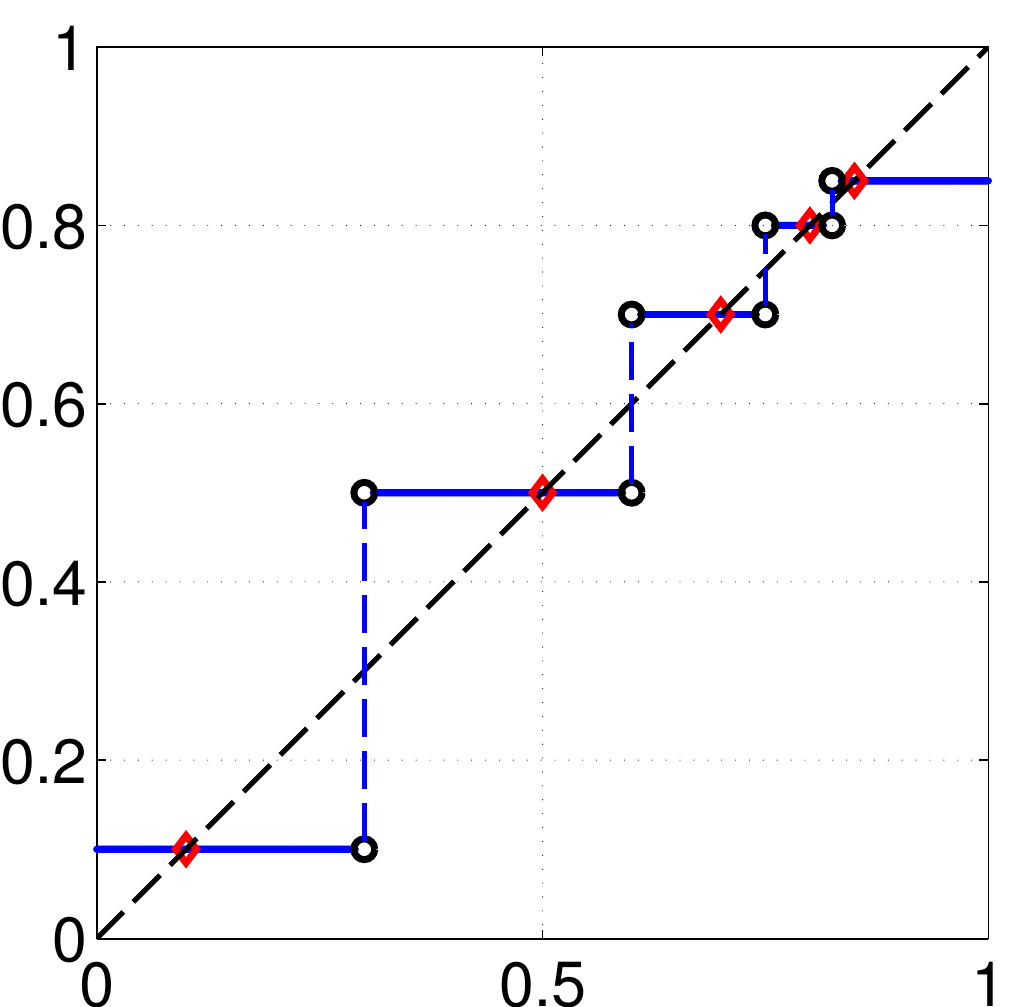}
\end{center}
~\vspace{-2.5em}
\caption{\em \small Examples of function $\phi_{\t,\lambda}(c)$ for different $\lambda$ values and the same set of $\t=[0.1, 0.5, 0.7, 0.8, 0.85]$ (\textcolor{red}{$\diamond$}). Left: soft-rounding ($\lambda=0.6$). Right: rounding ($\lambda=1.1$).}
~\vspace{-2.5em}
\label{fig:3}
\end{figure}

The proof of Theorem \ref{thm} is given in the supplementary material. In Fig.\ref{fig:3} we show examples of $\phi_{\t,\lambda}$ with different $\lambda$ values and the same set of $\t$. As can be seen from Theorem \ref{thm} and Fig.\ref{fig:3}, when $\lambda \geq 1$, $\phi_{\t,\lambda}$ maps $c \in [t_j,t_{j+1}]$ to the nearest of $t_j$ or $t_{j+1}$, \ie, it rounds $c$ to either $t_j$ or $t_{j+1}$. For $0 < \lambda <1$, $\phi_{\t,\lambda}$ keeps a zone $[t_j+ d_j , t_{j+1} - d_j]\subset[t_j,t_{j+1}]$, inside which $c$ is not rounded but undergoes a linear transform, the slope of which is determined by $\lambda$. As such, we term Eq.\eqref{eq:13} as {\em soft-rounding}, drawing an analogy from {\em soft-thresholding}, which is the proximal operator of $\ell_1$ norm~\cite{Donoho:1995:ST}. The soft-rounding operator ``softens'' the rounding operator, and has an interesting effect on its output: if $c$ is close to the endpoints of an interval, it will be rounded to the nearest endpoint.  However, if $c$ lies in the middle range of an interval, it is ``nudged'' towards the nearest $t$ value through a linear interpolation.



Note that through the ALM framework, regularizer $\Gamma_I$ and the resulting soft-rounding operator can readily work with most existing RLS image restoration methods. The two regularizers, $\Gamma_N$ and $\Gamma_I$, collaborate in a closed loop to recover the uncorrupted image. The former restores structures (edges and contours), and the latter encourages the restored image to have the desirable distinct pixel values. Also, typical restoration methods tend to smooth the image, and the soft-rounding operator, by enhancing the contrasts of the restored image, serves as a countermeasure to such over-smoothing.

Our method is advantageous to the simpler approach that first restores the image and then applies rounding to project the pixel values back to the target values. Since rounding is performed independent of and after the restoration step, it may destroy the structures recovered by restoration and undo its effect. This difference is demonstrated in the subsequent experiments in Section \ref{sec:4}.


\subsection{Obtaining Distinct Pixel Values}
\label{sec:34}

We first discuss several general ways to obtain distinct pixel values to be used in $\Gamma_I$, depending on the application scenario and domain knowledge. (1) Distinct pixel values can be extracted from external sources. For instance, text images captured under similar illumination conditions might vary in spatial layout and visual contents, but they show very close distinct values for text and background pixels respectively. (2) Distinct pixel values can be specified through user interaction. We first recover a clean image from the degraded observation with existing restoration methods, and then manually select dominant pixel values from various regions as distinct pixel values. (3) For specific problems, distinct pixel values can be estimated from the degraded observation. For images contaminated with small Gaussian noise, their marginal intensity histograms might be roughly estimated from the noisy observations following a 1D deconvolution approach~\cite{Wangmeng:2014:GHP}, from which we can extract distinct pixel values.


Apart from general discussions, we also provide a simple but effective method to estimate distinct pixel values from a degraded image. A natural idea is to first recover a clean image with standard image restoration methods (e.g., L0) and then use clustering methods such as K-Means to extract the cluster centers as distinct pixel values. However, the image degradation process blurs across different distinct pixel values and shrinks the distances between them. This effect might still be evident in the restored image, sometimes leading to inaccurate centers as shown Table~\ref{tbl:4}. Inspired by the local two-color model~\cite{Joshi:2009:IDD} and its recent extension~\cite{Lai:2015:NCLP},  we propose a simple two-step estimation method. In the first step, we extract a small patch around each pixel and estimate a local one-center or two-center model. If the intensity variance in the patch is below a threshold, the patch falls into the one-center model, and its mean pixel value is collected for the second global clustering step; otherwise the patch is  a two-center patch, and its two cluster centers are collected with a {\em normalized} two-class K-Means clustering method~\cite{Lai:2015:NCLP}. Compared to K-Means, normalized K-Means tries to stretch the distance between the two cluster centers and fight against the shrinkage effect due to image degradation. Then in the second step, we perform a global K-Median clustering on all the collected local center values to determine the final distinct pixel values. We provide the estimation results for three L0 deconvolution text and pattern image examples in Table~\ref{tbl:4}: the pixel values estimated by our method are very close to those of the clean images and are more accurate than the results produced by a single global clustering method such as K-Means.

\begin{table}
\footnotesize
\centering
\begin{tabular}{|c||c|c|c|}\hline
\multirow{2}{*}{Method} & \multicolumn{3}{|c|}{Examples}\\
\cline{2-4}
& Fig.~\ref{fig:teaser} & Fig.~\ref{fig:4} & Fig.~\ref{fig:7} (2nd Row) \\
\hline
True Values & $0,100,150,255$ & $26,217$ & $24,53,158,224,255$\\
\hline
K-Means & $18,98,156,245$ & $62,215$ & $14,55,164,221,253$\\
\hline
Our Method & $9,101,155,255$ & $32,217$ & $20,52,162,224,255$\\
\hline
\end{tabular}
~\vspace{-1.em}
\caption{\small \em The distinct pixel values estimated by K-Means and our method for three L0 deconvolution visual examples.}
~\vspace{-3.0em}
\label{tbl:4}
\end{table}

\section{Experiment}
\label{sec:4}
In this section, we combine the soft-rounding operator with several state-of-the-art RLS image restoration methods and evaluate its performance on a variety of images. We focus on the restoration of three typical categories of images with distinct pixel values, namely, text images with two pixel values, pattern images with multiple pixel values, and natural images with pixels of full range of $8$-bit depth. Furthermore, we also test our algorithm on other types of images that are of practical importance, such as barcodes, posters and license plates. In all experiments, we assume the distinct pixel values are known.

\subsection{Text Image Deconvolution}
\label{sec:4:1}

\begin{figure*}
\centering
\begin{subfigure}[t]{.3\textwidth}
\centering
\includegraphics[width=\textwidth]{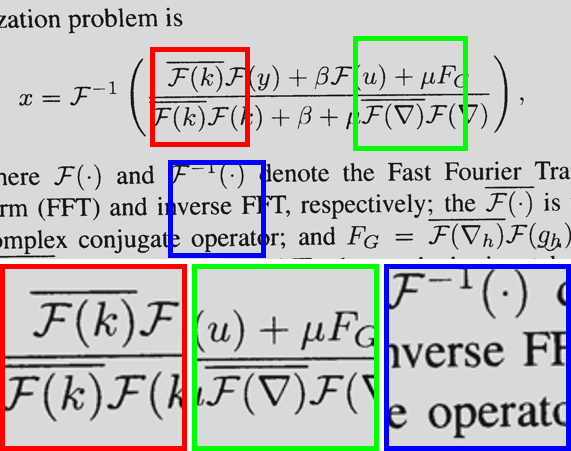}
\caption{\footnotesize original image}
\end{subfigure}
\begin{subfigure}[t]{.3\textwidth}
\centering
\includegraphics[width=\textwidth]{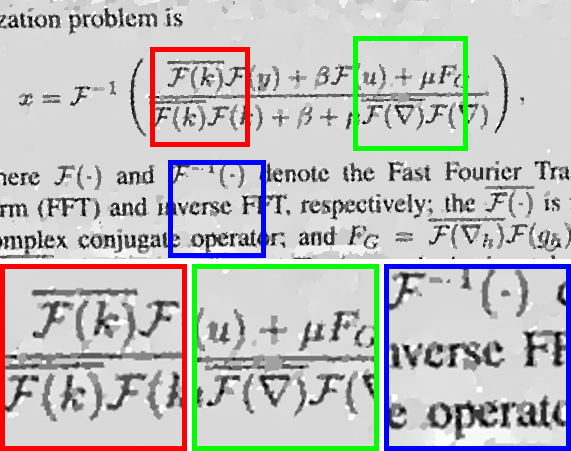}
\caption{\footnotesize L0 (PSNR = 20.83dB, SSIM = 0.86)}
\end{subfigure}
\begin{subfigure}[t]{.3\textwidth}
\centering
\includegraphics[width=\textwidth]{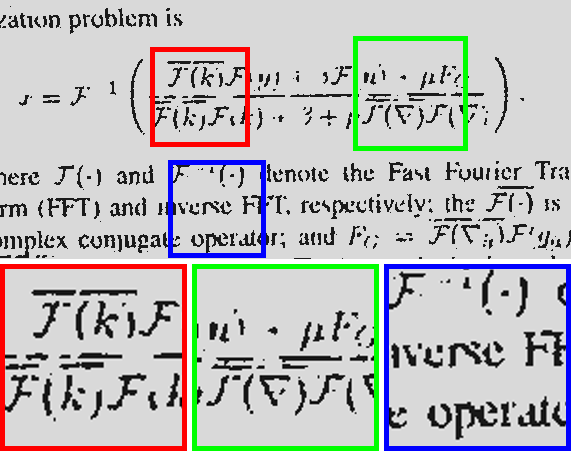}
\caption{\footnotesize L0+R (PSNR = 18.38dB, SSIM = 0.85)}
\end{subfigure}\\
\begin{subfigure}[t]{.3\textwidth}
\centering
\includegraphics[width=\textwidth]{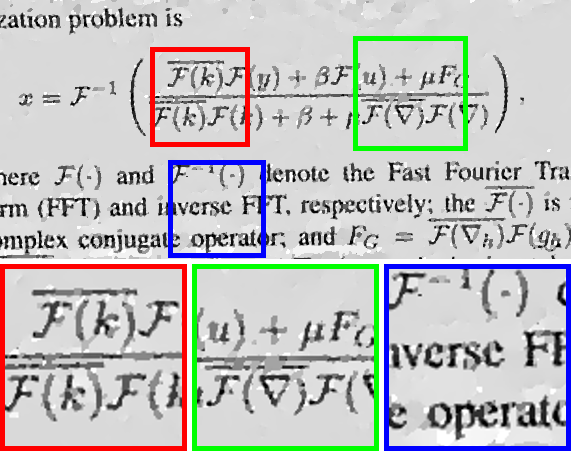}
\caption{\footnotesize L0+L (PSNR = 21.01dB, SSIM = 0.87)}
\end{subfigure}
\begin{subfigure}[t]{.3\textwidth}
\centering
\includegraphics[width=\textwidth]{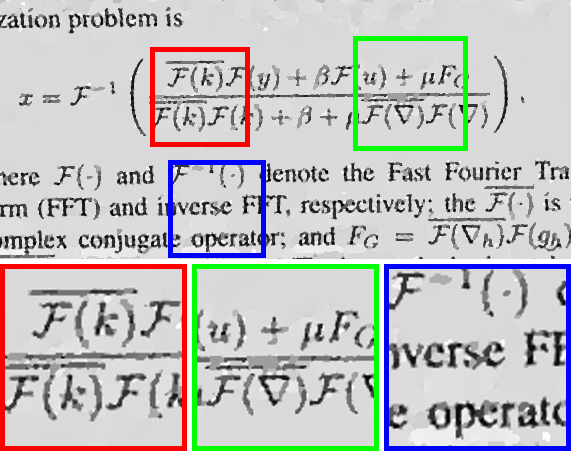}
\caption{\footnotesize L0+P (PSNR = 21.72dB, SSIM = 0.90)}
\end{subfigure}
\begin{subfigure}[t]{.3\textwidth}
\centering
\includegraphics[width=\textwidth]{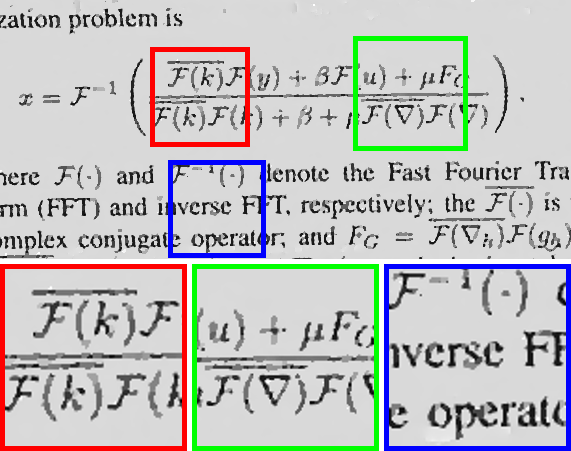}
\caption{\footnotesize L0+S (\textbf{PSNR = 21.88dB, SSIM = 0.92})}
\end{subfigure}
~\vspace{-1em}
\caption{\em \small Text Image Deconvolution Example. The original image is degraded with a $51\times51$ kernel and $1\%$ Gaussian noise. The values for text and background pixels are $t_1 = 26,t_2=217$ respectively. Details are best viewed on screen.}
\label{fig:4}
~\vspace{-2.5em}
\end{figure*}


Text images form an important but special class of images. Many text images originate from text documents and have two distinct pixel values for text and background pixels respectively. State-of-the-art non-blind restoration performance for text images was achieved with an RLS method with an $\ell_0$ regularizer in the gradient domain (subsequently denoted as L0 in Table \ref{tbl:1})~\cite{Pan:2014:DTI,Xu:2011:ISV,Xu:2013:UL0}. Assuming the pixel values for foreground and background pixels have been extracted from similar text images, we augment the $\ell_0$-based RLS restoration method with soft-rounding and denoted it as L0+S. We also compare the performance of L0 combined with three other pixel domain regularizers: a simple post-processing method which directly rounds the restored image to a binary image (denoted as L0+R), the fourth-order polynomial regularizer to incorporate binary intensity constraints~\cite{Zhang:2012:AAM} (denoted as L0+P) and the $\ell_0$ intensity regularizer from~\cite{Pan:2014:DTI} (denoted as L0+L). For fair comparison of each method, we adjust parameters and report the best overall performance.

We collect $20$ text images of different languages, equations and graphs. We use three kernel/noise settings to simulate the degradation process: (i) a $33\times33$ kernel and additive Gaussian noise with $3\%$ standard deviation; (ii) a $45\times45$ kernel and $2\%$ Gaussian noise; and (iii) a $51\times51$ kernel and $1\%$ Gaussian noise. The kernels are motion-blur kernels previously used in \cite{Levin:2007:IDC,Pan:2014:DTI}.

Table~\ref{tbl:1} reports the average peak-signal-to-noise (PSNR) and structural similarity index (SSIM)~\cite{Wang:2004:SSIM} results of the five methods in three kernel-noise settings. Compared to the original L0 method, the soft-rounding enhanced restoration method significantly improves the average PSNR by at least $0.5$dB and the average SSIM by at least $0.04$ for all the settings, which is more effective than the other two methods with intensity regularizers~(L0+L, L0+P). The simple rounding method (L0+R) actually lowers the performance for most settings. These quantitative results demonstrate the effectiveness of our method.

\begin{table} [h]
\footnotesize
\centering
\begin{tabular}{|c||c|c|c|c|c|}\hline
\multirow{2}{*}{Kernel/Noise} & \multicolumn{5}{|c|}{Avg. PSNR (in dB)}\\
\cline{2-6}
& L0 & L0+R & L0+L & L0+P & L0+S\\
\hline
$33\times33$+$3\%$ & $19.46$ & $16.75$ & $19.71$ & $19.91$ & $\mathbf{20.03}$\\
\hline
$45\times45$+$2\%$& $18.20$ & $15.71$ & $18.47$ & $18.57$ & $\mathbf{18.86} $\\
\hline
$51\times51$+$1\%$ & $18.03$ & $15.95$ & $18.23$ & $18.93$ & $\mathbf{19.16}$\\
\hline
\hline
\multirow{2}{*}{Kernel/Noise} &\multicolumn{5}{|c|}{Avg. SSIM}\\
\cline{2-6}
& L0 & L0+R & L0+L & L0+P & L0+S\\
\hline
$33\times33$+$3\%$ & $0.86$ & $0.84$ & $0.87$ & $0.88$ & $\mathbf{0.90}$\\
\hline
$45\times45$+$2\%$ & $0.82$ & $0.81$ & $0.83$ & $0.86$ & $\mathbf{0.88}$\\
\hline
$51\times51$+$1\%$ & $0.78$ & $0.83$ & $0.79$ & $0.83$ & $\mathbf{0.86}$\\
\hline
\end{tabular}
~\vspace{-1em}
\caption{\small \em The average PSNR and SSIM results of the $20$ text images in three kernel-noise settings.}
~\vspace{-2.5em}
\label{tbl:1}
\end{table}

We provide a visual example of kernel/noise setting (iii) in Figure~\ref{fig:4}, which contains an equation and some English text. As shown by the example, the original L0 method produces pixel values that do not exist in the original text image, leading to many visible artifacts. A direct rounding to the binary image~(L0+R) removes out-of-range pixel values, but it also destroys recovered structures, which lowers both visual quality and quantitative measures. Using the $\ell_0$ regularizer in pixel domain (L0+L) slightly improves over the L0 method in PSNR and SSIM, but visual artifacts are still visible in the restored image. On the other hand, regularizers specifically enforcing distinct pixel values including the method of \cite{Zhang:2012:AAM} (L0+P) and our regularizer (L0+S) can more effectively incorporate the binary intensity constraint. Our method recovers more structures and suppresses more intensity artifacts with the soft-rounding operator, leading to the best visual quality and quantitative results. More examples can be found in the supplemental material.

\begin{figure*}
\centering
\begin{subfigure}[t]{.16\textwidth}
\centering
\includegraphics[width=\textwidth]{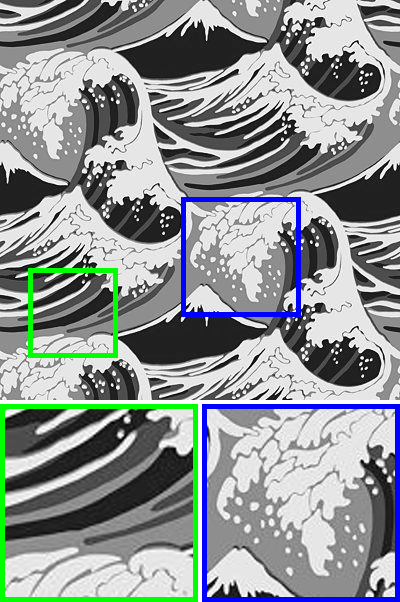}
\caption{\footnotesize original image}
\end{subfigure}
\begin{subfigure}[t]{.16\textwidth}
\centering
\includegraphics[width=\textwidth]{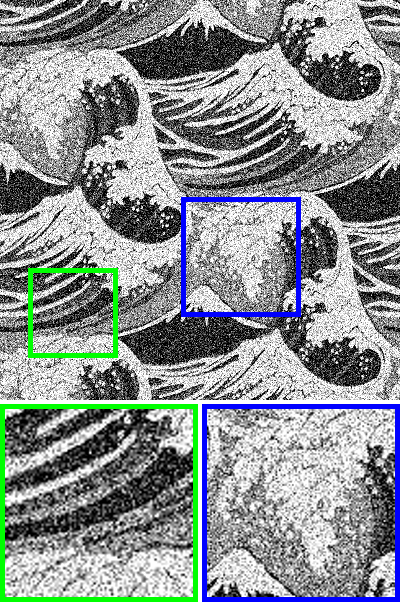}
\caption{\footnotesize noisy observation}
\end{subfigure}
\begin{subfigure}[t]{.16\textwidth}
\centering
\includegraphics[width=\textwidth]{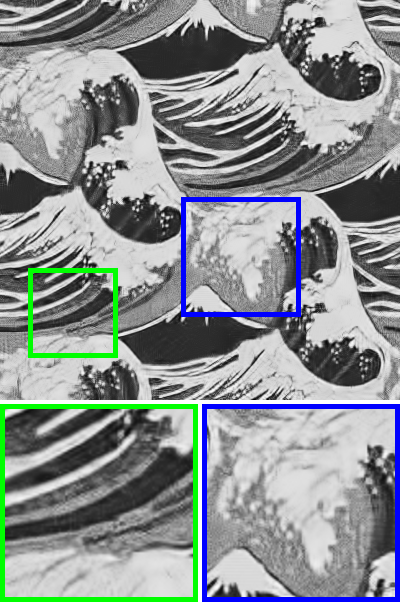}
\caption{\footnotesize \centering BM3D \newline PSNR = 19.97dB \newline SSIM = 0.69}
\end{subfigure}
\begin{subfigure}[t]{.16\textwidth}
\centering
\includegraphics[width=\textwidth]{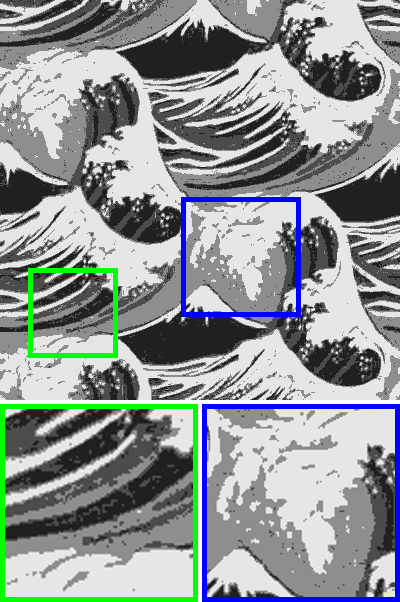}
\caption{\footnotesize \centering BM3D+R \newline PSNR = 19.82dB \newline SSIM = 0.71}
\end{subfigure}
\begin{subfigure}[t]{.16\textwidth}
\centering
\includegraphics[width=\textwidth]{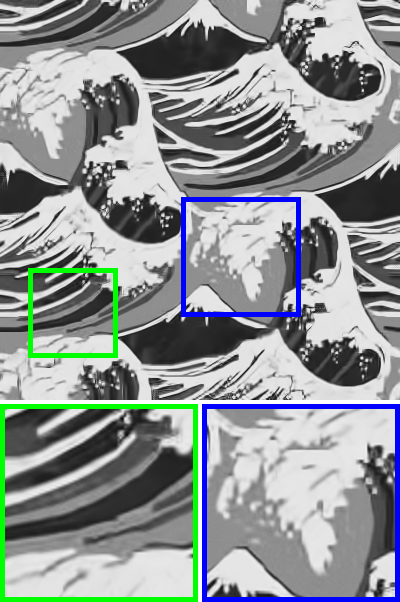}
\caption{\footnotesize \centering BM3D+S \textbf{\newline PSNR = 20.93dB \newline SSIM = 0.76}}
\end{subfigure}\\
\setcounter{subfigure}{0}
\begin{subfigure}[t]{.16\textwidth}
\centering
\includegraphics[width=\textwidth]{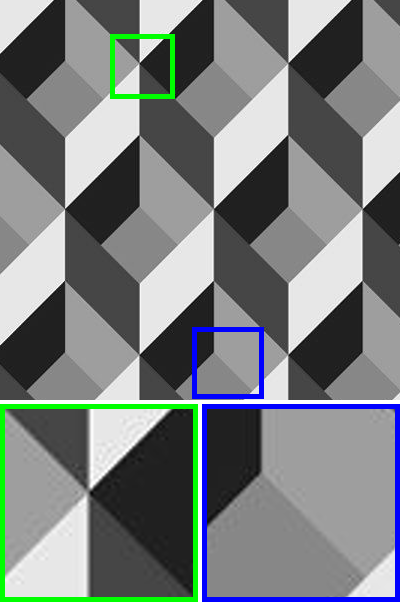}
\caption{\footnotesize original image}
\end{subfigure}
\begin{subfigure}[t]{.16\textwidth}
\centering
\includegraphics[width=\textwidth]{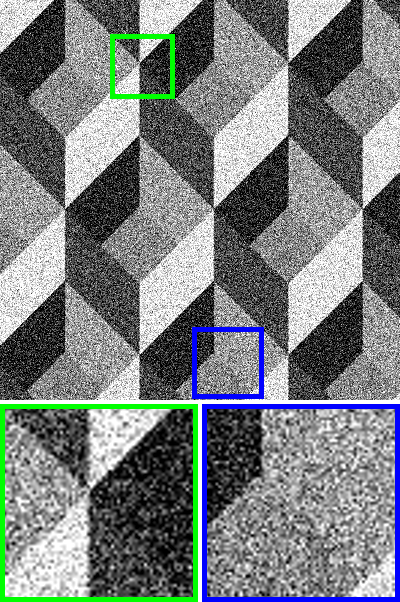}
\caption{\footnotesize noisy observation}
\end{subfigure}
\begin{subfigure}[t]{.16\textwidth}
\centering
\includegraphics[width=\textwidth]{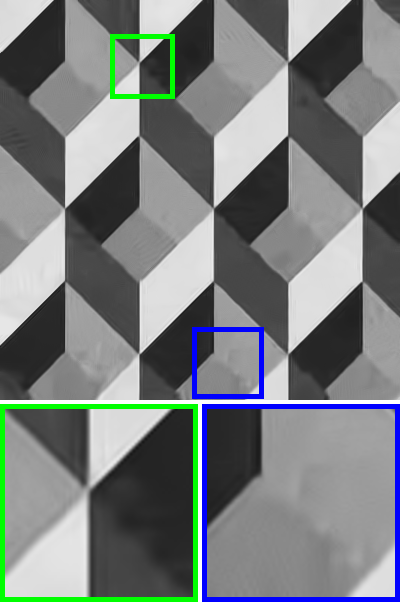}
\caption{\footnotesize \centering BM3D \newline PSNR = 29.65dB \newline SSIM = 0.93}
\end{subfigure}
\begin{subfigure}[t]{.16\textwidth}
\centering
\includegraphics[width=\textwidth]{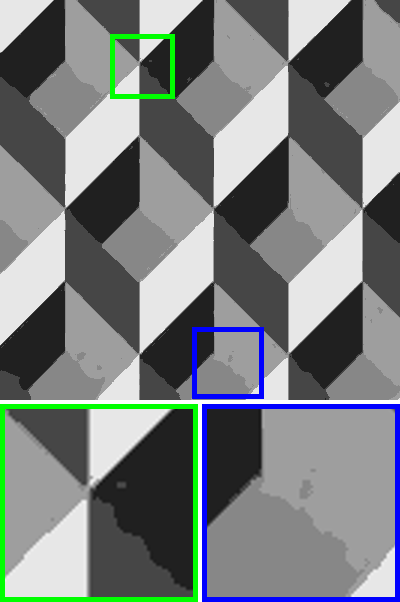}
\caption{\footnotesize \centering BM3D+R \newline PSNR = 32.02dB \newline SSIM = 0.94}
\end{subfigure}
\begin{subfigure}[t]{.16\textwidth}
\centering
\includegraphics[width=\textwidth]{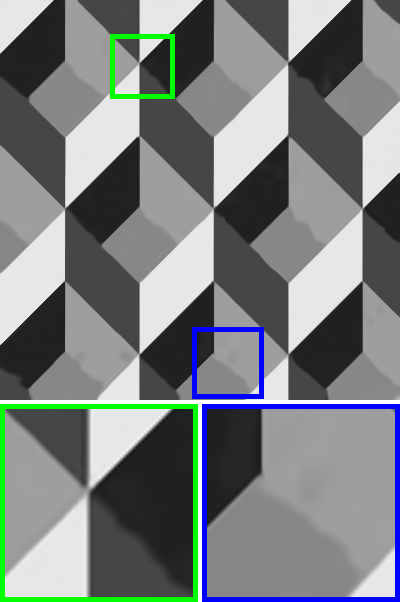}
\caption{\footnotesize \centering BM3D+S \textbf{\newline PSNR = 34.35dB \newline SSIM = 0.96}}
\end{subfigure}
~\vspace{-1em}
\caption{\em \small Pattern Image Denoising Examples. Top row: this pattern image is degraded with $25\%$ Gaussian noise. The distinct pixel values are $32, 76, 142, 230$ respectively. Bottom row: this pattern image is degraded with $20\%$ Gaussian noise. The distinct pixel values are $32, 70, 135, 158, 231$ respectively. Details are best viewed on screen.}\label{fig:6}
~\vspace{-2.5em}
\end{figure*}

\subsection{Pattern Image Restoration}
\label{sec:4:2}

Pattern images are an important class of images which can be frequently found from human-made objects, artistic designs and paintings. An important property of pattern images is that their visual contents are represented with a few distinct pixel values~\cite{Mei:2015:UniHIST}. We test our method to remove noise from pattern images. In our experiments, we collect $10$ pattern image examples and degrade them with three high Gaussian noise levels ($15\%$, $20\%$ and $25\%$), such that the restoration performance
depends heavily on the regularizers. We adopt the BM3D denoiser as the baseline method~\cite{Dabov:2007:BM3D} and combine it with the soft-rounding regularizer (denoted as BM3D+S). The numerical solution of BM3D+S can be found in the supplemental material. For comparison, we include a post-processing method which directly rounds the BM3D denoised image pixels to nearest known pixel values (denoted as BM3D+R). Note that previous binary intensity regularizers~\cite{Pan:2014:DTI,Zhang:2012:AAM} can not be easily extended to handle more than two distinct pixel values.

\begin{table}
\footnotesize
\centering
\begin{tabular}{|c||c|c|c|}\hline
\multirow{2}{*}{Noise} & \multicolumn{3}{|c|}{Avg. PSNR (in dB)}\\
\cline{2-4}
& BM3D & BM3D+R & BM3D+S\\
\hline
$15\%$ & $29.77$ & $29.75$ & $\mathbf{31.13}$\\
\hline
$20\%$& $26.71$ & $27.21$ & $\mathbf{28.67}$\\
\hline
$25\%$ & $24.59$ & $25.21$ & $\mathbf{25.96}$\\
\hline
\hline
\multirow{2}{*}{Noise} & \multicolumn{3}{|c|}{Avg. SSIM}\\
\cline{2-4}
& BM3D & BM3D+R & BM3D+S\\
\hline
$15\%$ & $0.92$ & $0.91$ & $\mathbf{0.94}$\\
\hline
$20\%$ & $0.89$ & $0.89$ & $\mathbf{0.92}$\\
\hline
$25\%$ & $0.86$ & $0.85$ & $\mathbf{0.88}$\\
\hline
\end{tabular}
~\vspace{-1.em}
\caption{\small \em The average PSNR and SSIM results of the $10$ pattern images in three noise settings.}
~\vspace{-3.5em}
\label{tbl:2}
\end{table}

\begin{figure*}
\centering
\begin{subfigure}[t]{.17\textwidth}
\includegraphics[width=\textwidth]{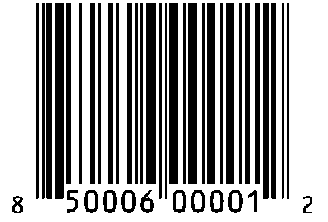}
\caption{\footnotesize original image}
\end{subfigure}
\begin{subfigure}[t]{.17\textwidth}
\includegraphics[width=\textwidth]{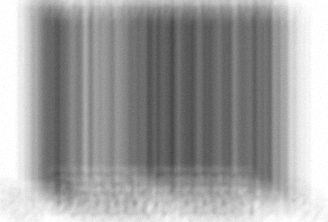}
\caption{\footnotesize degraded image}
\end{subfigure}
\begin{subfigure}[t]{.17\textwidth}
\includegraphics[width=\textwidth]{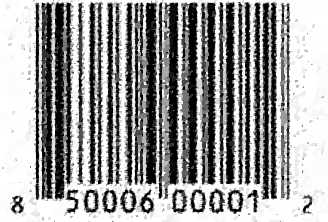}
\caption{\footnotesize \centering L0 \newline PSNR = 13.18dB \newline SSIM = 0.70}
\end{subfigure}
\begin{subfigure}[t]{.17\textwidth}
\includegraphics[width=\textwidth]{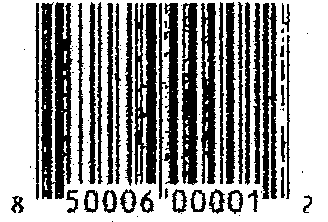}
\caption{\footnotesize \centering L0+R \newline PSNR = 12.50dB \newline SSIM = 0.80}
\end{subfigure}
\begin{subfigure}[t]{.17\textwidth}
\includegraphics[width=\textwidth]{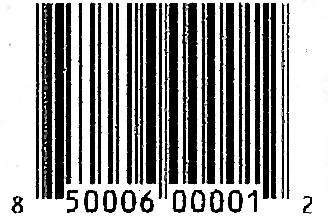}
\caption{\footnotesize \centering L0+S \textbf{\newline PSNR = 21.02dB \newline SSIM = 0.97}}
\end{subfigure}\\
\setcounter{subfigure}{0}
\begin{subfigure}[t]{.17\textwidth}
\includegraphics[width=\textwidth]{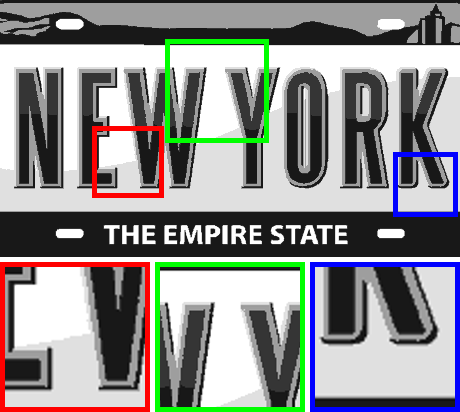}
\caption{\footnotesize original image}
\end{subfigure}
\begin{subfigure}[t]{.17\textwidth}
\includegraphics[width=\textwidth]{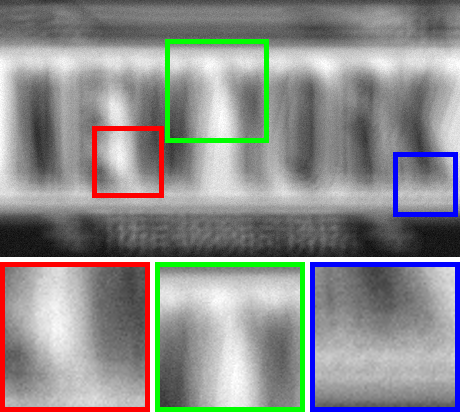}
\caption{\footnotesize degraded image}
\end{subfigure}
\begin{subfigure}[t]{.17\textwidth}
\includegraphics[width=\textwidth]{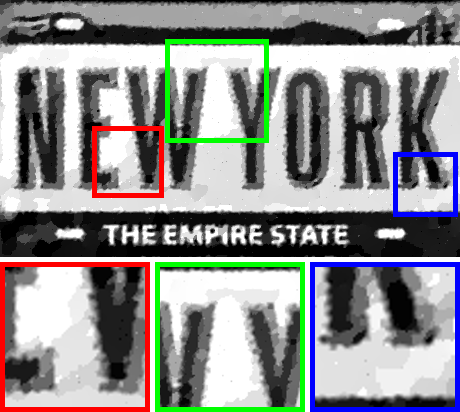}
\caption{\footnotesize \centering L0 \newline PSNR = 18.69dB \newline SSIM = 0.67}
\end{subfigure}
\begin{subfigure}[t]{.17\textwidth}
\includegraphics[width=\textwidth]{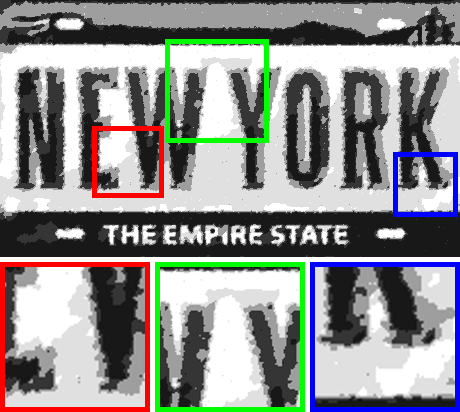}
\caption{\footnotesize \centering L0+R \newline PSNR = 17.87dB \newline SSIM = 0.72}
\end{subfigure}
\begin{subfigure}[t]{.17\textwidth}
\includegraphics[width=\textwidth]{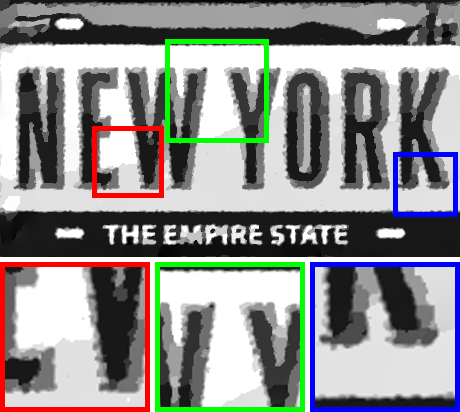}
\caption{\footnotesize \centering L0+S \textbf{\newline PSNR = 19.13dB \newline SSIM = 0.78}}
\end{subfigure}
~\vspace{-1em}
\caption{\em \small More L0 Deconvolution Examples. Top row: the 1D barcode image is degraded with a $65\times65$ kernel and $1\%$ Gaussian noise. The distinct pixel values for barcode and background pixels are $t_1 = 0,t_2=255$ respectively. Bottom row: the license plate image is degraded with a $55\times55$ kernel and $3\%$ Gaussian noise. The distinct pixel values are $24, 53, 158, 224, 255$ respectively. Details are best viewed on screen.}\label{fig:7}
~\vspace{-2em}
\end{figure*}

Table~\ref{tbl:2} reports the average PSNR and SSIM results over 10 pattern images for the three noise levels. Compared to BM3D, our method (BM3D+S) improves the average PSNR by at least $1.3$dB and the average SSIM by at least $0.02$ for the three cases. The simple rounding method (BM3D+R) sometimes leads to a deterioration either in PSNR or in SSIM. We provide two visual examples with different noise levels are in Figure~\ref{fig:6}. As these results show, the original BM3D method tends to introduce out-of-the-range pixel values and blurred boundaries. The post rounding method (BM3D+R) improves the image contrast with known intensity information, but it also introduces visible errors on region boundaries. On the other hand, our method improves both visual quality and quantitative results with better contrast and image structures.

We further present several deconvolution examples of practical pattern images in Figure~\ref{fig:7}. We use the L0 deconvolution method as the baseline method and combine the distinct pixel value information with the L0+R and L0+S method. As in the case of denoising, our method (L0+S) significantly improves the visual quality and the quantitative results, while the direct rounding method (L0+R) introduces extra artifacts and removes useful structures in the original images. Please refer to the supplemental material for more examples.

\subsection{Natural Image Deconvolution}
\label{sec:4:3}

We randomly select 200 natural images from the BSDS500 dataset~\cite{Martin:2001:BSDS300} and test them with the same kernel/noise settings used in Sec.~\ref{sec:4:1}. For natural image deconvolution, we choose the total variation method (denoted as TV)~\cite{Rudin:1992:TV} as our baseline method. We integrate the integer intensity constraint into the TV method using soft-rounding (denoted as TV+S) and compare it to the post-rounding method (denoted as TV+R).



Our experiments show that compared to TV, TV+S improves the average PSNR by $0.005$dB, $0.005$dB and $0.025$dB for the three cases, respectively, while TV+R slightly degrades the performance by about $0.002$dB for each case. Even though the average contribution of the constraint on integer pixel values seems marginal, a detailed performance breakdown can shed more light on its effect. Specifically, we report the PSNR difference between TV+S and TV on the $200$ images in Table~\ref{tbl:3}: for all the three kernel/noise settings, TV+S improves TV on most images; and for at least $11\%$ of the images, the improvement obtained with TV+S over TV is at least $0.01$dB; on the other hand, it decreases TV performance significantly ($>0.01$dB) only on one image example from all the cases.
\smallskip

\noindent\textbf{Summary}. As all the experimental results show, even though the constraint on pixel values is simple, effectively incorporating it in image restoration often improves the visual quality and the quantitative results for a variety of images and various restoration methods. Furthermore, for special classes of natural images that have distinct pixel values, it leads to considerable performance improvement.


\begin{table}
\footnotesize
\centering
\begin{tabular}{|c||c|c||c|c|}\hline
\multirow{2}{*}{Kernel/Noise} & \multicolumn{4}{|c|}{\# of Images with PSNR Difference $D$}\\
\cline{2-5}
& $>0$ & $<0$ & $>0.01$ & $<-0.01$\\
\hline
$33\times33$+$3\%$ & $\mathbf{129}$ & $71$ & $\mathbf{24}$& $0$\\
\hline
$45\times45$+$2\%$& $\mathbf{126}$ & $74$ & $\mathbf{22}$ & $0$\\
\hline
$51\times51$+$1\%$ & $\mathbf{132}$ & $68$ & $\mathbf{68}$& $1$\\
\hline
\end{tabular}
~\vspace{-1.em}
\caption{\small \em Quantitative comparison of TV+S and TV on $200$ BSDS500 images for the three kernel/noise settings. The PSNR difference is defined as $D = {PSNR}_{TV+S}-{PSNR}_{TV}$.}
~\vspace{-2.5em}
\label{tbl:3}
\end{table}


\section{Conclusion}
\label{sec:5}
In this work, we describe a new regularizer to augment current image restoration algorithms when the original images have known distinct pixel values. Our regularizer is in the form of a  function and can be efficiently implemented with a {\em soft-rounding} operation. This regularizer can be readily incorporated into most existing image restoration methods and our experiments on restoration of binary text images, pattern images with multiple pixel values and natural images show that its incorporation leads to considerable performance improvements to the state-of-the-art methods.

There are a few directions we would like to further improve the current work. Currently, our method relies on the knowledge of the distinct pixel values in the original image. Another scenario is when we only know the exact number of distinct pixel values but not the values themselves. One subsequent study is to augment our algorithm so that it can also simultaneously estimate the distinct pixel values and restore the corrupted image. Furthermore, discriminative image restoration methods~\cite{Schmidt:2014:SF,Schmidt:2013:DNBD} seem to be an effective alternative to the generative methods, and it is also of interest to combine our regularizer with such methods. 

{\small
\bibliographystyle{ieee}
\bibliography{references}
}

\end{document}